\documentclass{article}

\usepackage[T1]{fontenc}
\usepackage{subcaption} 
\usepackage{booktabs}

\usepackage{amssymb}
\usepackage{amsmath}

\usepackage{hyperref}

\usepackage{cite}

\usepackage{graphicx}

\usepackage{authblk}

\usepackage[small]{titlesec}
\begin{document}
\title{\vspace{-2cm}Automatic Gradient Estimation for Calibrating Crowd Models with Discrete Decision Making}
\date{\vspace{-5ex}}

\author{Philipp Andelfinger, Justin~N.~Kreikemeyer\thanks{Email: \{philipp.andelfinger,justin.kreikemeyer\}@uni-rostock.de}}
\affil{Institute for Visual and Analytic Computing,\\University of Rostock, Germany}

\maketitle

\begin{abstract}

\noindent Recently proposed gradient estimators enable gradient descent over stochastic programs with discrete jumps in the response surface, which are not covered by automatic differentiation (AD) alone.
Although these estimators' capability to guide a swift local search has been shown for certain problems, their applicability to models relevant to real-world applications remains largely unexplored.
As the gradients governing the choice in candidate solutions are calculated from sampled simulation trajectories, the optimization procedure bears similarities to metaheuristics such as particle swarm optimization, which puts the focus on the different methods' calibration progress per function evaluation.
Here, we consider the calibration of force-based crowd evacuation models based on the popular Social Force model augmented by discrete decision making.
After studying the ability of an AD-based estimator for branching programs to capture the simulation's rugged response surface, calibration problems are tackled using gradient descent and two metaheuristics.
As our main insights, we find 1) that the estimation's fidelity benefits from disregarding jumps of large magnitude inherent to the Social Force model, and 2) that the common problem of calibration by adjusting a simulation input distribution obviates the need for AD across the Social Force calculations, allowing gradient descent to excel.

\end{abstract}

\section{Introduction}

Agent-based crowd models are widely used in urban planning~\cite{motieyan2018agent,gonzalez2021agent}, to study disease spread~\cite{hackl2019epidemic} or to optimize strategies for emergency evacuations~\cite{chen2008agent,yin2014agent,kasereka2018agent}.
In contrast to coarse-grained models that consider groups of people in aggregate, agent-based models operate on the microscopic level, giving each agent its own state, perception of its environment, and decision making.
To generate meaningful simulation results, the calibration of the model parameters to empirical data, e.g., gathered from video footage, is a crucial prerequisite.

Often, metaheuristics such as genetic algorithms or particle swarm optimization are applied~\cite{wolinski2014parameter,voloshin2015optimization}, which permit a straightforward parallelization to tackle the computational cost of evaluating microscopic simulations across high-dimensional parameter spaces.
Surrogate models generated by sampling the simulation response can support a faster calibration but may require enormous numbers of samples to capture an original model's dynamics~\cite{pietzsch2020metamodels}.

However, all these methods typically operate on black-box observations of the simulation output.
If it is possible to also determine its partial derivatives wrt.~the parameters, local search strategies based on gradient descent can steer the calibration toward a local optimum.
Calibration using Bayesian inference~\cite{bode2020parameter,godel2022bayesian}, which also offers uncertainty information instead of sheer point estimates, could particularly benefit from the resulting increase in sampling efficiency~\cite{cranmer2020frontier}.
Unfortunately, in the presence of models of discrete decision making, agent-based simulations form stochastic functions with discrete jumps.
For these functions, simple averaging over derivatives determined using the established methods and tools for automatic differentiation (AD)~\cite{griewank2008evaluating,margossian2019review} yields biased estimates~\cite{gong1987smoothed}.

Recently, there has been renewed interest in gradient estimation over stochastic functions with discrete jumps.
Rooted in infinitesimal perturbation analysis, this line of research has produced AD-based estimators tailored to specific domains~\cite{chopra2021deepabm,son2022differentiable,andelfinger2023towards}, for programs involving random sampling from discrete probability distributions~\cite{arya2022automatic}, and for general imperative programs~\cite{kreikemeyer2023smoothing}.
The recent generic estimators combine pathwise derivatives with the contributions of jumps, the former being determined using traditional AD, and the latter based on a priori knowledge of the distributions or density estimations.
As an AD-free alternative, modern finite differences estimators compute gradient estimates from series of function evaluations on stochastically perturbed inputs~\cite{nesterov2017random}.
These recent estimators' reliance on sampling raises the question whether the function evaluations permitted by a time budget are better spent on obtaining gradient estimates, or to directly evaluate a set of candidate solutions as part of a metaheuristic.

Here, we explore the suitability of gradient descent for calibrating crowd evacuation models based on Treiber's popular Social Force model~\cite{helbing1995social} augmented by discrete decisions.
We assess an AD-based and a stochastic finite differences-based estimator~\cite{nesterov2017random,kreikemeyer2023smoothing} compared to a genetic algorithm and particle swarm optimization.
Our main contributions are threefold:

\begin{itemize}
\item We present an alternative derivation of our gradient estimator DiscoGrad Gradient Oracle~\cite{kreikemeyer2023smoothing} starting from the concept of stratified derivatives.
\item We study the fidelity of sampling-based gradient estimates over an evacuation scenario with continuous or discrete objective.
\item The calibration progress is evaluated for three problems, one being a distribution fitting problem over a 20-dimensional parameter space, showing that this problem class permits fast convergence via gradient descent.
\end{itemize}

The remainder of the paper is structured as follows:
In Section~\ref{sec:background_and_related_work}, we briefly introduce methods for (automatic) differentiation across discrete jumps.
In Section~\ref{sec:dgo}, we introduce the existing AD-based gradient estimator DGO.
Section~\ref{sec:crowd_model_and_scenarios} describes the considered simulation model and scenarios.
In Section~\ref{sec:experiments}, we present our experiment results and discuss their implications.
Section~\ref{sec:conclusions} provides an interpretation of our results and concludes the paper.

\section{Background and Related Work}
\label{sec:background_and_related_work}

Simulation models of crowd dynamics typically combine low-level models of pedestrian movement with 1) models of discrete decisions, e.g., for path planning, and 2) stochastic components accounting for uncertainty and variability in initial conditions and pedestrian behaviors.
In effect, the models thus take the form of stochastic functions $\mathcal{P}\colon\mathbb{R}^n\to\mathbb{R}$ involving discrete jumps, which poses challenges to traditional gradient estimation methods.
In the following, we briefly introduce the existing work on gradient estimation across discontinuities and concrete estimators for this purpose.

To begin, we briefly recapitulate the widely employed concept of \emph{automatic differentiation}~(AD) \cite{griewank2008evaluating,margossian2019review}.
This method views the execution of $\mathcal{P}$ as a composition of operators $\mathcal{P}_1\circ\mathcal{P}_2\circ\,\cdots$.
By repeatedly applying the chain rule, the partial derivatives wrt.\ the inputs can be determined from the intermediate derivatives and values at the operators $\mathcal{P}_i$.
Implementations can be grouped into \emph{reverse} and \emph{forward} modes.
Whereas the former implements AD as a second (reverse) pass over previously stored intermediate values retrieving one row of the Jacobian per pass, the latter propagates derivatives through the forward execution retrieving one column per pass.
The partial derivatives obtained this way are \emph{pathwise} in the sense that they only capture the operation sequence of a single program execution, disregarding alternative branches.

When estimating gradients of stochastic programs, we are typically interested in the partial derivatives of the expected value of $\mathcal{P}$ wrt.\ a parameter vector $\theta$:
\begin{equation}
  \frac{\partial}{\partial\theta}\mathbb{E}\left[\mathcal{P}(\theta)\right] 
\label{eq:smooth_derivative}
\end{equation}

\noindent As is common in the literature \cite{seyer2023differentiable,gong1987smoothed}, we introduce an additional parameter $\omega$.
This allows the explicit consideration of the stochasticity of $\mathcal{P}$, so that $\mathcal{P}(\omega;\theta)$ refers to a specific realization of the stochastic function, e.g., as determined by a pseudo-random generator's seed.
Without loss of generality, we sometimes consider only $n=1$, while $n>1$ follows directly from separately calculating the partial derivative for each dimension.
One important case occurs if $\mathcal{P}$ exhibits discontinuities (jumps), whose positions depend on $\theta$ and/or $\omega$.
In the last decades, several methods have been developed to deal with this situation, an overview of which is given in~\cite{fu2006chapter}.

Among the earliest is the \emph{infinitesimal perturbation analysis} (IPA) estimator~\cite{ho1983new}.
Relying on an interchange of the differentiation and expectation operators in Eq.~\ref{eq:smooth_derivative} it can be computed by averaging over pathwise derivatives.
However, this estimator is biased for discontinuous $\mathcal{P}$, as then the requirements for the interchange of operators are not satisfied.
To still account for jumps, \emph{smoothed perturbation analysis} (SPA)~\cite{gong1987smoothed} employs a method inspired by Conditional Monte Carlo.
Based on the law of total expectation, $\mathbb{E}[\mathcal{P}]$ is calculated as $\mathbb{E}[\mathbb{E}[\mathcal{P}|z]]$ for some tailored characterization $z$ of $\mathcal{P}$'s execution.
If $z$ is chosen correctly, this allows the use of pathwise derivatives as in IPA.

Recently, inspired by the success of AD on deterministic programs, automatic methods to calculate Eq.~\ref{eq:smooth_derivative} gained new interest.
The \emph{StochasticAD} estimator~\cite{arya2022automatic} builds on SPA and AD to allow the automatic differentiation of programs sampling from discrete parametric distributions. 
Other recent publications employ \emph{interpolation}, replacing discontinuous operators $\mathcal{P}_i$ in $\mathcal{P}$ with continuous approximations \cite{andelfinger2023towards,christodoulou2023differentiable}, and \emph{abstract interpretation} \cite{chaudhuri2010smooth}, symbolically propagating distributions through $\mathcal{P}$ to smooth over discontinuities. 

Another approach to the calculation of Eq.~\ref{eq:smooth_derivative} are black-box estimators like REINFORCE \cite{williams1992simple} and \emph{randomized finite-differences} schemes.
A notable candidate from the latter category is proposed in \cite{nesterov2017random} building on \cite[Chapter 3.4]{polyak1987introduction}, which we adopt here as follows under the name Polyak Gradient Oracle (PGO):
\begin{equation}
  \nabla \mathcal{P}(\mathbf{\theta}) \approx \textstyle\sum_{s=1}^{S} (\mathcal{P}(\mathbf{\theta}+\sigma\mathbf{u},\omega_s) - \mathcal{P}(\mathbf{\theta},\omega))\sigma^{-1}\mathbf{u} / S,
\label{eq:pgo_estimator}
\end{equation}

\noindent where $\mathbf{u}$ is a vector of i.i.d. standard normal variates and $\sigma$ a ``smoothing factor''.
By introducing random perturbations on $\theta$, this estimator can provide a full gradient estimate from one sample.
Note that introducing such perturbations is possible (or even required) with many estimators, allowing their application to deterministic programs with discontinuities.

\section{DiscoGrad Gradient Oracle}
\label{sec:dgo}

In this publication, we evaluate practical applications of the recently proposed DiscoGrad Gradient Oracle (DGO) \cite{kreikemeyer2023smoothing}.
The following provides an alternative derivation starting from the abstract concept of the ``stratified derivative'' from~\cite{seyer2023differentiable}.
The latter is constructed around the concept of a \emph{critical event} $A$, which occurs if $|\mathcal{P}(\omega;\theta + \epsilon / 2) - \mathcal{P}(\omega;\theta - \epsilon / 2)| > B|\epsilon|$ for some bound $B > 0$, i.e., when a jump is observed in an $\epsilon$-neighborhood around $\theta$.
Then, it holds that
\begin{equation}
\frac{\partial \mathbb{E} \left[\mathcal{P}(\theta) \right] }{\partial \theta} = \mathbb{E} \left[ \frac{\partial}{\partial \theta} \mathcal{P}(\theta) \right] + \mathbb{E} \left[{\Delta}_\mathcal{P} \right] p'_\theta.
\label{eq:stratified_derivative}
\end{equation}

\noindent Here, $\Delta_\mathcal{P}$ denotes the distribution of the jump's magnitudes conditioned on $A$, and $p'_\theta$ is the \emph{critical rate}, defined as $\lim_{\epsilon \downarrow 0} \frac{1}{\epsilon} \mathbb{P}(A)$.
The DGO estimates the above for the special case of imperative programs with conditional branches.

Let us consider a program $\mathcal{P}$ with scalar input and output, including a single branch of the form ``if G($\theta$) < d``, with $d$ a constant and the value of $\mathcal{P}$ depending on the path taken.
Defining $C(\theta):=G(\theta) - d$, the branching condition can be rewritten as $C(\theta) < 0$.
A realization of $C$'s sign indicates the chosen branch.

The term $\mathbb{E} \left[ \frac{\partial}{\partial \theta} \mathcal{P}(\theta) \right]$ in Eq.~\ref{eq:stratified_derivative} can be trivially estimated by sampling pathwise derivatives using AD.
Considering $p'_\theta$, we first note that $\mathbb{P}(|\mathcal{P}(\omega;\theta + \epsilon / 2) - \mathcal{P}(\omega;\theta - \epsilon / 2)| > B|\epsilon|) = \mathbb{P}( C(\omega;\theta + \epsilon / 2) \cdot C(\omega;\theta - \epsilon / 2) < 0)$.
This is the probability of a sign change in an $\epsilon$-neighborhood around $\theta$.
We assume all jumps to originate from branches of the above form, which still allows many other discontinuous functions like the minimum or absolute value to be expressed.

For $\epsilon \rightarrow 0$, we arrive at $p'_\theta = f_{C(\theta)}(0) \mathbb{E} \left[ \frac{\partial}{\partial \theta}C(\theta) \right]$, where $f_{C(\theta)}$ is the probability density function of $C(\theta)$. 
DGO estimates $p'_\theta$ by gathering realizations $C(\omega_s;\theta)$ and calculating a density estimation, which is evaluated at the origin.
$\mathbb{E} \left[ \frac{\partial}{\partial \theta}C(\theta) \right]$ as well as $\mathbb{E} \left[{\Delta}_\mathcal{P} \right]$ are estimated using realizations of $C(\theta)$ close to $0$.

Estimating gradients across programs with several branches requires additional considerations.
In the presence of sequential branches, a branch condition's distribution can depend on whether previous branches have been taken.
Hence, the density estimation must distinguish the control flow path along which a branch is reached.
We uniquely identify each branch $b$ by the path along which it is encountered, corresponding to the sequence of condition signs at all previous branches.
Now, we can express DGO for programs with $\mathfrak{B} \in \mathbb{N}$ branches:
\begin{equation}
  \frac{\partial \mathbb{E} \left[\mathcal{P}(\theta) \right] }{\partial \theta} =
  \lim \limits_{S \rightarrow \infty} \frac{1}{S}\sum_{s=1}^{S} \frac{\partial \mathcal{P}(\omega_s;\theta)}{\partial \theta}  +\kern-0.5em \\
  \sum_{b=1}^{\mathfrak{B}} \left(\mathcal{P}(\omega^{\texttt{+}}_{b};\theta) - \mathcal{P}(\omega^{\texttt{-}}_{b};\theta)\right) \lambda_b \hat{f}_{C_b(\theta)}(0)\frac{\partial C^\epsilon_b}{\partial \theta}
\label{eq:dgo_estimator}
\end{equation}

\noindent where $\omega^{\texttt{+}}_{b}$ and $\omega^{\texttt{-}}_{b}$ select the samples $s$ corresponding to the positive and negative realizations of $C_b(\theta)$ closest to the branching point, $\lambda_b$ is the proportion of samples that encountered the branch, $\hat{f}_{C_b(\theta)}$ is an estimate of $C_b(\theta)$'s probability density function, and $\frac{\partial}{\partial\theta}C^\epsilon_b$ is the partial derivative of the condition near the branching point.

We note that, in contrast to PGO (cf.~Section~\ref{sec:background_and_related_work}), whose calculation of directional derivatives relies on perturbations of the parameters, DGO can operate on an original program without introducing external stochasticity.
However, by reducing the ruggedness of the objective function, smoothing via external perturbations can contribute to faster convergence of gradient descent.

An implementation of DGO exists as part of the DiscoGrad tool~\cite{kreikemeyer2023smoothing}, which permits the differentiation across a subset of C++ programs with conditional branches.
The implementation is available publicly\footnote{\url{https://github.com/philipp-andelfinger/DiscoGrad}}.

\section{Crowd Model and Scenarios}
\label{sec:crowd_model_and_scenarios}

\begin{figure}[t!]
\centering
\includegraphics[width=0.42\textwidth,angle=90]{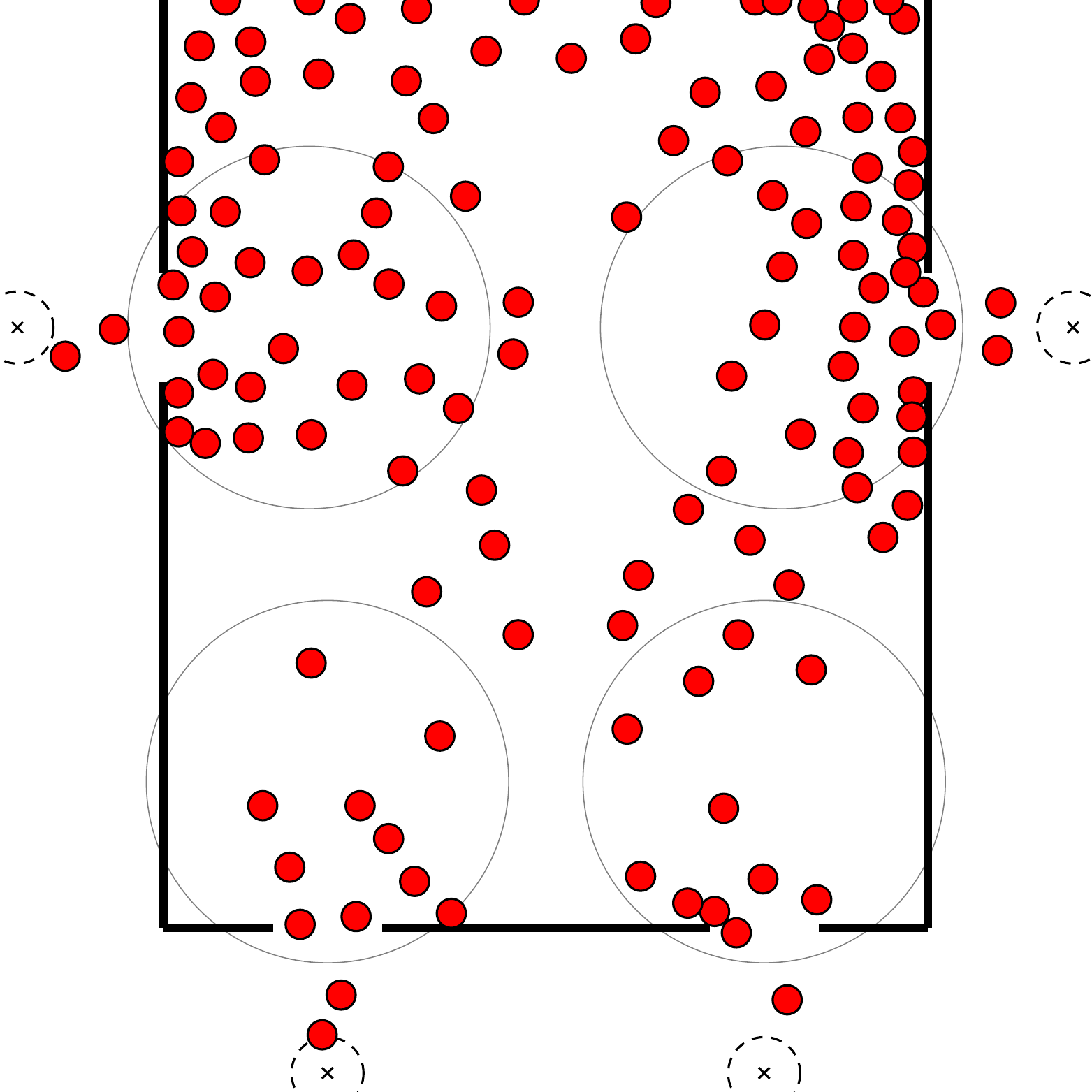}
\caption{Scenario for calibrating exit selection coefficients. The crowd enters from the left-hand side, aiming to evacuate by reaching the circular waypoints (dashed circles), each agent periodically reconsidering the targeted exit by weighing its distance against the number of agents in its vicinity (gray circles).}
\label{fig:scenario}
\end{figure}

Our experiments build on the typical constituents of evacuation studies: a force-based model of crowd mobility in a two-dimensional continuous space and per-pedestrian discrete decision making.
The mobility is modeled using Treiber's popular Social Force model~\cite{helbing1995social}, in which a person's effective acceleration vector is calculated as a sum of three forces.
The \emph{internal} force reflects a person $i$'s intention to move in a straight line towards its goal location in direction $e_i^0$ with desired velocity $v_i^0$, $v_i(t)$ being the current velocity. 
The adaptation time is scaled according to a characteristic time $\tau_i$.
\emph{Interaction} forces $f_ {ij}$ between person $i$ and each other person $j$ in his or her vicinity exert a repellent effect, reflecting avoidance maneuvers and maintenance of personal distance.
Finally, \emph{obstacle} forces $f_{iW}$ repel the person from any nearby wall $W$, leading to the overall force equation for a person $i$ with mass $m_i$ and scaling coefficients $w_1$, $w_2$, $w_3$:
$$m_i \frac{dv_i}{dt} = w_1 m_i \frac{v_i^0 e_i^0(t) - v_i(t)}{\tau_i} + w_2 \sum_{j \ne i} f_{ij} + w_3 \sum_W f_{iW}$$

We consider two scenarios, the first representing a bottleneck in an evacuation situation, similar to~\cite{godel2022bayesian}.
The simulation space is a 30 $\times$ 30m square separated in the center by a single wall with a door 4m in width.
As in~\cite{godel2022bayesian}, we calibrate the weight coefficients that determine the strengths of the forces experienced by pedestrians aiming to pass through the door.
The output of the simulation to be calibrated is either the pedestrians' average horizontal position or the number of pedestrians evacuated after 20s of simulation time.

The second scenario (cf.~Figure~\ref{fig:scenario}) combines low-level mobility via the Social Force model with discrete exit selections as in existing work such as~\cite{wang2022modeling}.
A crowd comprised of 1\,000 pedestrians gradually enters the scenario from the left-hand side, aiming to exit the building via any of the four available doors, each 3m wide.
Each pedestrian selects its target door by weighing its distance against the congestion level at the door as measured by the number of agents nearby.
To be able to react to changing circumstances, each agent reconsiders the previous decision every 15s.
The weight coefficient underlying the decision is drawn from a probability distribution supplied as model parameters in the form of a histogram ranging in 20 steps from 0.1, where the decision is dominated by the congestion level, to 1.0, where the decision is made solely based on distance.
The simulation output to be minimized is the Wasserstein distance of the histogram of observed evacuation times to a reference histogram after a warm-up time of 100s, spanning 20 steps from 10s to 75s.

The scenarios differ fundamentally in their implications for gradient estimation.
While the first scenario involves explicit conditional branches only in the counting of evacuations, discrete jumps are created by the Social Force model itself, making this a challenging scenario for DGO.
In contrast, the calibration of the second scenario leads to gradients entirely defined by conditional branches, which prevents their estimation via AD alone but is well-suited for DGO.

\section{Experiments}
\label{sec:experiments}

The goal of our experiments is to determine whether gradient descent using sampling-based gradient estimators can outperform genetic algorithms and particle swarm optimization in the calibration of evacuation models.
We approach this objective by first studying the degree to which the estimators are able to capture Social Force's dynamics.
We then turn to the calibration problems and carry out hyperparameter sweeps in order to shed light on the relative performance of the different optimization methods.

The simulation models were implemented in C++ within DiscoGrad~\cite{kreikemeyer2023smoothing}, closely following PEDSIM\footnote{\url{https://github.com/chgloor/pedsim}} for the Social Force model and its parametrization.
As the genetic algorithm implementation, we used pyeasyga~\footnote{\url{https://github.com/remiomosowon/pyeasyga}}, and for the particle swarm optimization we employ the pyswarms library~\cite{miranda2018pyswarms}.
All simulations use Leapfrog integration with a time step of 0.1s.
The calibration experiments were carried out on two identical machines, each equipped with an AMD EPYC 9754 processor with 256 threads and 768GB RAM, running Ubuntu 22.04.4 LTS, each machine executing at most 256 calibration runs in parallel.

\begin{figure*}[t!]
\centering
\begin{minipage}{0.49\textwidth}
\subfloat[3 agents, internal weight $w_0$.]{\includegraphics[height=2.42cm]{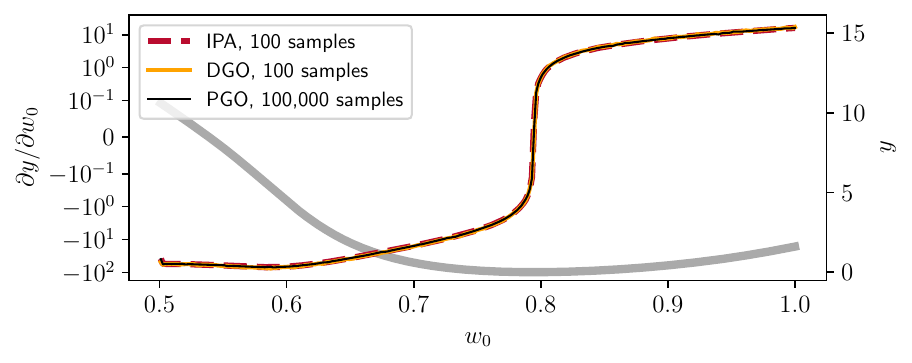}}

\hspace{0.15cm}\subfloat[3 agents, interaction weight $w_1$.]{\includegraphics[height=2.42cm]{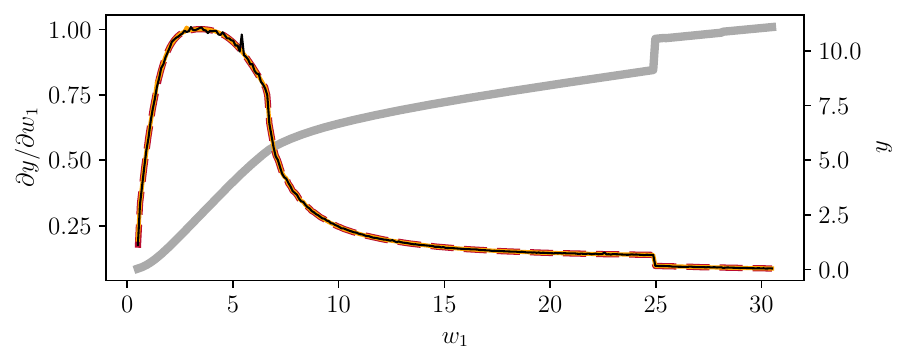}}

\subfloat[3 agents, obstacle weight $w_2$.]{\includegraphics[height=2.42cm]{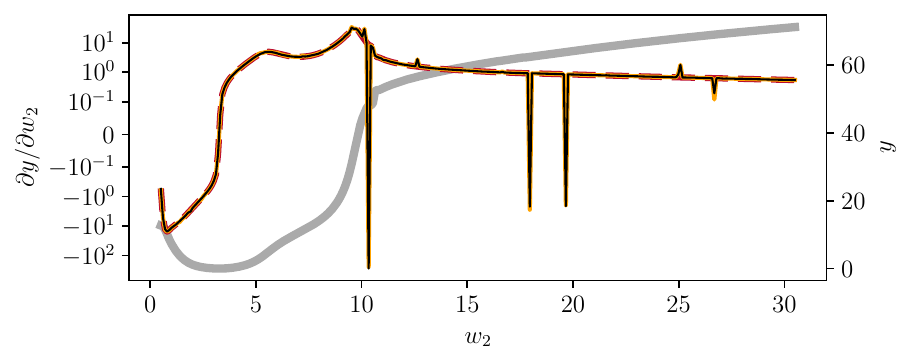}}
\end{minipage}
\hfill
\begin{minipage}{0.49\textwidth}
\subfloat[10 agents, internal weight $w_0$.]{\includegraphics[height=2.42cm]{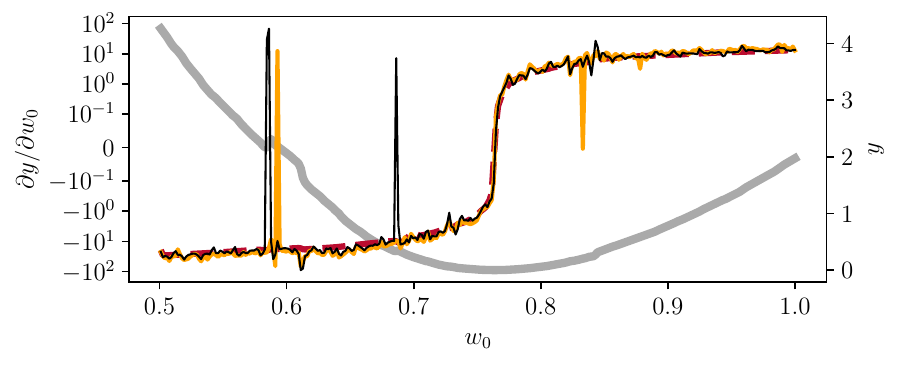}}

\hspace{0.21cm}\subfloat[10 agents, interaction weight $w_1$.]{\includegraphics[height=2.42cm]{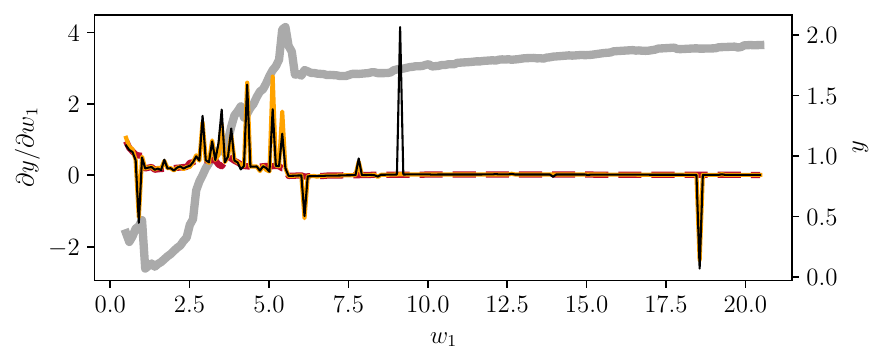}}

\subfloat[10 agents, obstacle weight $w_2$.]{\includegraphics[height=2.42cm]{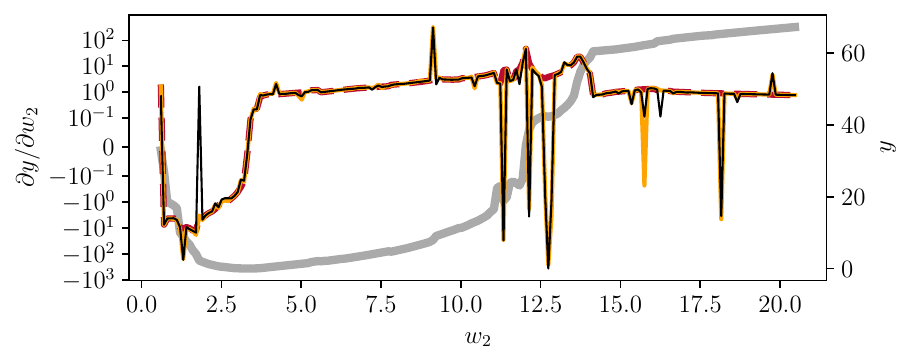}}
\end{minipage}

\caption{Derivative estimates with respect to the weight of the three types of forces in scenarios populated with 3 agents (a-c) and 10 agents (d-f) with the fit in the agents' final coordinates' as output and $\sigma = 0.001$. PGO with a large number of samples is used as the reference. DGO captures most of the derivatives' spikes, whereas the IPA estimate only reflects the general curvature.}
\label{fig:y_pos_gradients}
\vspace{0.2cm}
\end{figure*}

\subsection{Automatically Differentiating the Social Force Model}

While the Social Force model itself does not call for explicit conditional branches, slight perturbations in the parameters can cause large discrete jumps in acceleration.
For instance, the interaction force increases exponentially as the distance between two agents decreases, and its direction depends discretely on the distance and angle difference between two agents.
If the distance between two agents is small, minor changes in parameters can cause a change in direction and thus extreme changes in forces.
While AD alone correctly determines the gradient at a given point of the parameter space, it cannot capture such jumps.
DGO can treat these jumps as explicit conditional branches.
However, its estimation relies on sampled derivatives of intermediate branch conditions (cf.~Section~\ref{sec:dgo}), which may suffer from high variance when pairs of agents come in close mutual proximity.

To assess the differences in the AD-based and black-box estimators' ability to estimate gradients of crowd simulations, we carried out a parameter sweep across the weight coefficients $w_0$, $w_1$, $w_2$ that govern the intensity of the internal, interaction, and obstacles forces for the single-exit evacuation scenario.
For competitiveness with optimization procedures other than gradient descent, we are particularly interested in the gradients' fidelity with small numbers of samples.
Thus, we study the estimation error when varying the number of samples in comparison to reference gradients calculated from $100\,000$ function evaluations via PGO, which delivers unbiased estimates of the smoothed gradient.

Figure~\ref{fig:y_pos_gradients} shows the partial derivatives wrt.~each of the coefficients being varied separately in 300 steps while keeping the others fixed at values of $0.6$, $5.5$, and $5.5$, respectively.
The grey curves show the simulation output, which is the squared error in the average final agent positions compared to the pre-defined reference.
For comparability to PGO, the input is perturbed by Gaussian noise with $\sigma = 0.001$.
For the partial derivative estimates, we evaluate averaging across plain AD gradients (IPA) and DGO against the reference produced by PGO.
With 3 agents, both AD-based estimators closely track the derivatives' curvature.
In (b), we can see that the jump in the simulation output at about 24.5 is not visible in the derivative at the chosen resolution along the $w_1$ axis.
Importantly, we observe that the IPA curve does not follow the sharp downward spikes in $\partial y / \partial w_2$, while they are accurately captured by DGO.
The results with 10 agents follow a similar trend, but due the increased number of force calculations, the simulation output becomes substantially more rugged, resulting in noisier estimates using DGO.

\begin{figure*}[t!]
\centering
\begin{minipage}{0.49\textwidth}

\subfloat[3 agents, $\sigma = 0.01$.]{\includegraphics[height=2.42cm]{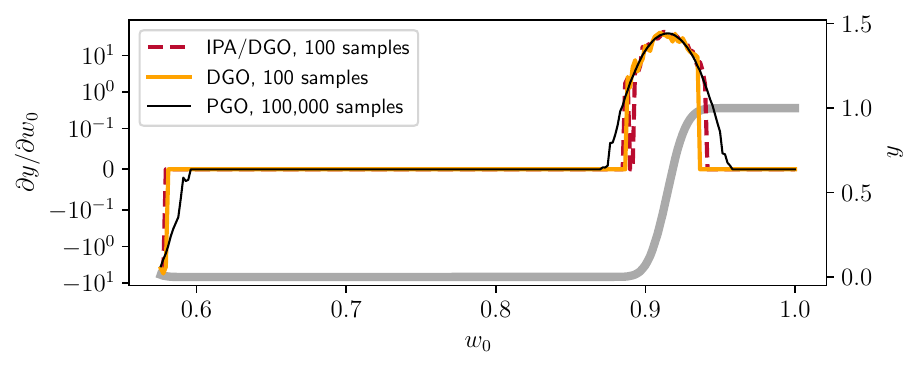}}

\hspace{0.11cm}\subfloat[3 agents, $\sigma = 0.1$.]{\includegraphics[height=2.42cm]{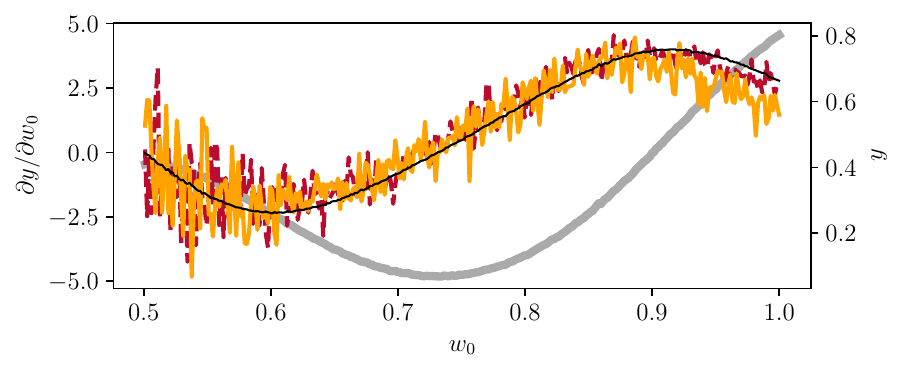}}
\end{minipage}
\hfill
\begin{minipage}{0.49\textwidth}

\hspace{0.02cm}\subfloat[10 agents, $\sigma = 0.01$.]{\includegraphics[height=2.42cm]{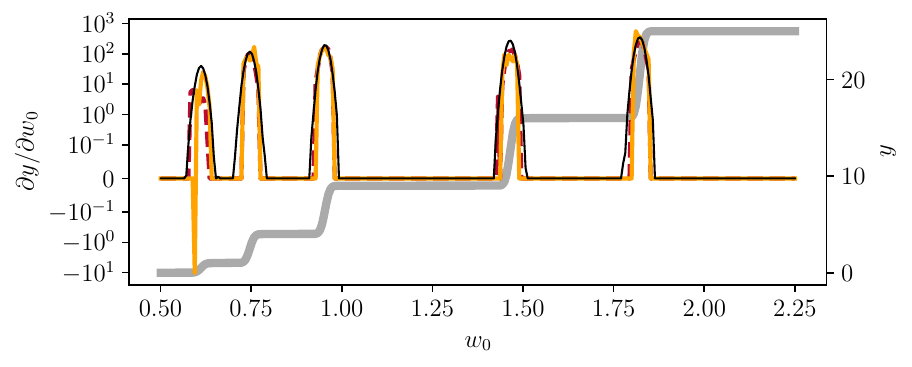}}

\subfloat[10 agents, $\sigma = 0.1$.]{\includegraphics[height=2.42cm]{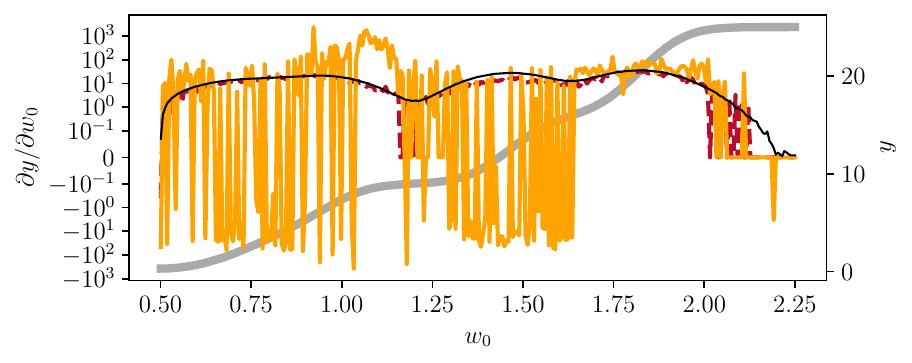}}
\end{minipage}

\caption{Derivative estimates with respect to the weight of the internal force, the simulation output being the fit in the number of evacuations. In the larger scenario, DGO's estimates suffer from substantial noise as jumps in the mobility derivatives translate to biased and high-variance derivative estimates of the simulation output. When ignoring jumps in the Social Force model (IPA/DGO), estimates observe some bias but capture the trends well.}
\label{fig:num_evac_gradients}
\vspace{0.2cm}
\end{figure*}

Figure~\ref{fig:num_evac_gradients} assesses the same scenario focusing on $w_0$ after changing the output to the squared error in the number of evacuations compared to a reference value.
This entirely discrete objective is smoothed only by the parameter's perturbations, whose standard deviation we set to $0.01$ and $0.1$, observing the expected increase in smoothness with the larger value.
In this problem, the simulation output is gathered by counting the number of agents that have passed the exit, which in the model's source code translates to a series of conditional branches on the agent positions.
Considering DGO, the estimation of the branches being taken and their effects on the overall derivatives are now subject to any noise in the positions' partial derivatives, leading to extremely noisy and often inaccurate derivative estimates particularly with 10 agents.

In the previous scenario, we have seen that the IPA estimates reflect the main curvature of the derivatives.
However, since IPA alone yields zero-valued derivatives with this discrete objective, it cannot be applied here.
Instead, we combine IPA and DGO by disregarding any jumps in Social Force while still accounting for the effects of branches using DGO.
In this combination, while some noise and slight deviations from the reference are observed, the tendencies of the reference are reflected much more accurately with 100 function evaluations.

\begin{figure}[t!]
\centering
\subfloat[3 agents.]{\includegraphics[width=0.245\textwidth]{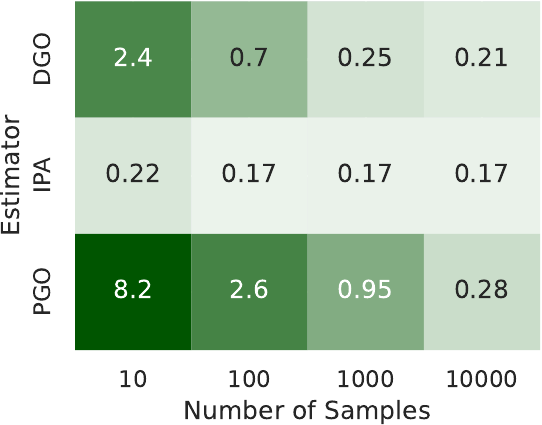}}\hspace{2cm}
\subfloat[10 agents.]{\includegraphics[width=0.245\textwidth]{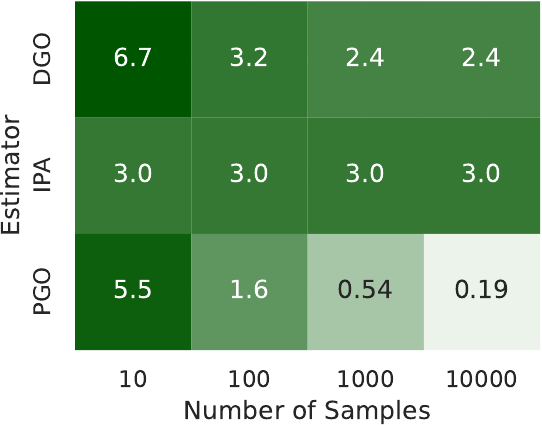}}
\caption{MAE of gradients wrt.~$w_0$ fitting the agent coordinates, $\sigma = 0.001$.}
\label{fig:fidelity_heatmap_y_pos}
\vspace{-0.2cm}
\end{figure}

\begin{figure}[t!]
\centering
\subfloat[3 agents.]{\includegraphics[width=0.245\textwidth]{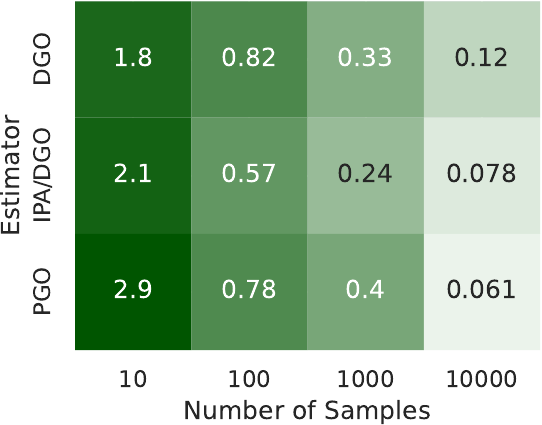}}\hspace{2cm}
\subfloat[10 agents.]{\includegraphics[width=0.245\textwidth]{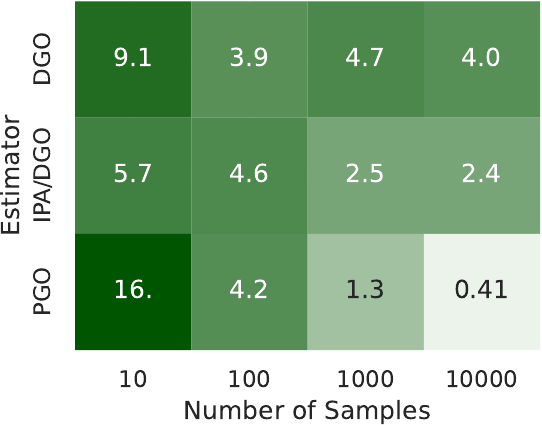}}
\caption{MAE of gradients wrt.~$w_0$ fitting the evacuation count, $\sigma = 0.001$.}
\label{fig:fidelity_heatmap_num_evac}
\end{figure}

To quantify the gradient estimates' fidelity, we consider the mean absolute error compared to PGO with $100\,000$ function evaluations for $w_0$.
Figure~\ref{fig:fidelity_heatmap_y_pos} supports the previous observations:
As the source of IPA's deviation from the reference is its inability to capture jumps, its estimates do not improve with additional samples, while the estimates with 10 samples observe a similar error compared to the other estimators.
In contrast, DGO improves somewhat with more evaluations but consistently outperforms PGO only in the smaller scenario.
PGO reliably approaches the reference when increasing the sample count.
However, we note that at low sample counts, the AD-based estimators are highly competitive.

The results for the same scenario with the discrete objective are shown in Figure~\ref{fig:fidelity_heatmap_num_evac}.
Here, similar error levels are observed for all estimators with 3 agents, while with 10 agents, PGO is superior at 1\,000 function evaluations and beyond.
Again, the AD-based estimators are competitive up to 100 evaluations.

\subsection{Calibrating Force Coefficients}

We now compare the practical capabilities of the different gradient estimators, PSO and GA.
To achieve a reasonably fair comparison among the calibration progress, we carried out a sweep across a range of sensible hyperparameters, with 10 microreplications of the program and 20 macroreplications of each hyperparameter configuration (cf.~Table~\ref{tab:hyperparams}).

\setlength{\tabcolsep}{1pt}
\begin{table}[b!]
  \centering
\scriptsize
  \begin{tabular}{lr}
    \multicolumn{2}{c}{\textbf{DGO, PGO, and IPA}} \\
    \toprule
    samples & 1, 10, 50\\
    \midrule
    $\sigma$ & 0, 0.01, 0.1, 0.5, 1.0 \\
    \midrule
    lr & 0.01, 0.1, 0.5, 1.0\\
    \bottomrule
  \end{tabular}\hfill
  \begin{tabular}{lr}
    \multicolumn{2}{c}{\textbf{PSO}} \\
    \toprule
    particles & 10, 50 \\
    \midrule
    $c_1$, $c_2$, w & 10 LHC samples\\
    \midrule
    neighbors & 3, 6, all \\
    \bottomrule
  \end{tabular}\hfill
  \begin{tabular}{lr}
    \multicolumn{2}{c}{\textbf{GA}} \\
    \toprule
    population & 10, 50 \\
    \midrule
    elitism & yes, no \\
    \midrule
    mutation & $x\texttt{=}\mathcal{U}(0,10)$, $x\texttt{+=}\mathcal{N}(0,0.1)$ \\
    \bottomrule
  \end{tabular}
  \caption{Hyperparameter sampling ranges. Legend: $\sigma$ sample size, lr learning rate, $(c_1,c_2,w)$ social, cognitive, and inertia parameters, LHC latin hyper cube.}
  \label{tab:hyperparams}
\vspace{-.4cm}
\end{table}

\begin{figure*}[t!]
\subfloat[Progress over wall time.]{\includegraphics[height=2.6cm]{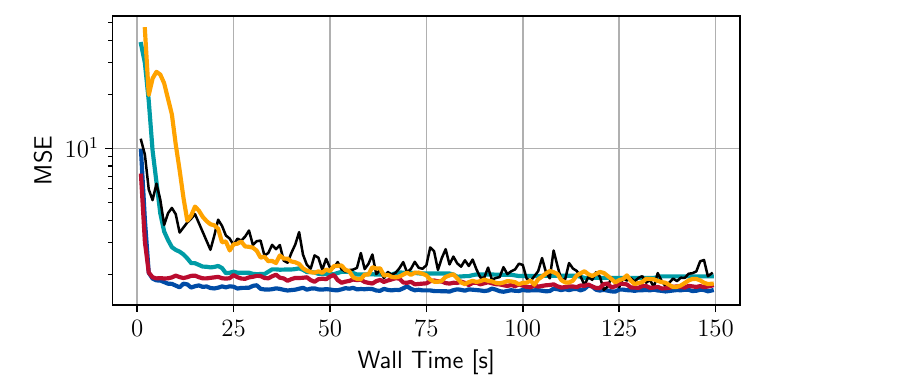}}\hfill
\subfloat[Progress over function evaluations.]{\includegraphics[height=2.6cm]{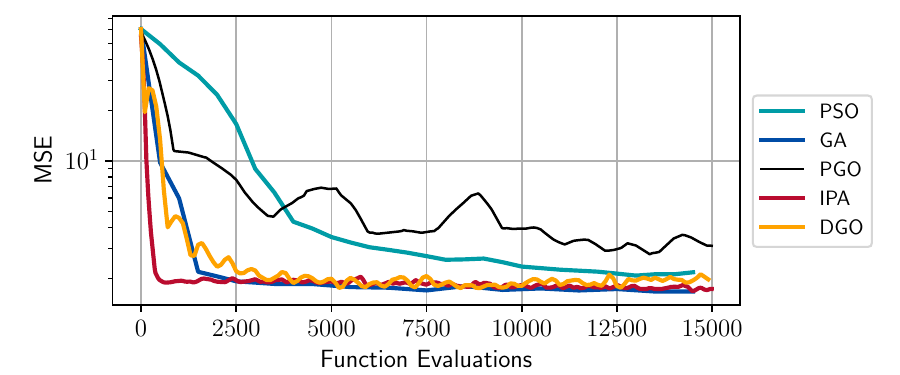}}
\caption{Calibration progress over steps, wall time and function evaluations during calibration of the final agent coordinates via the Social Force weights.}
\label{fig:opt_progress_y_pos}
\vspace{-.4cm}
\end{figure*}

\begin{figure*}[t!]
\subfloat[Progress over wall time.]{\includegraphics[height=2.6cm]{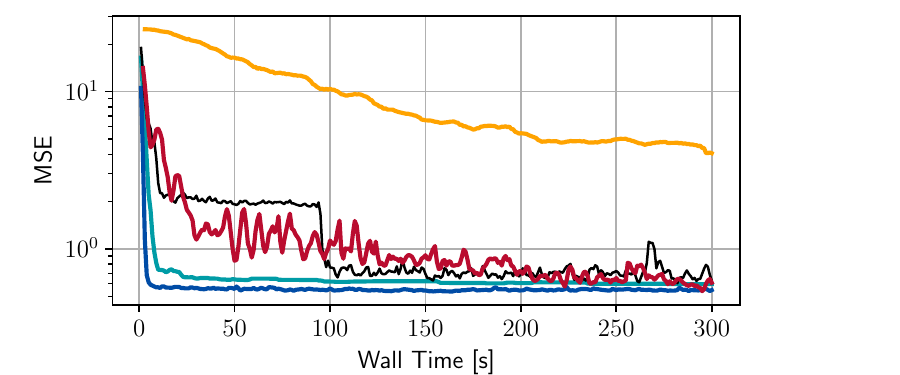}}\hfill
\subfloat[Progress over function evaluations.]{\includegraphics[height=2.6cm]{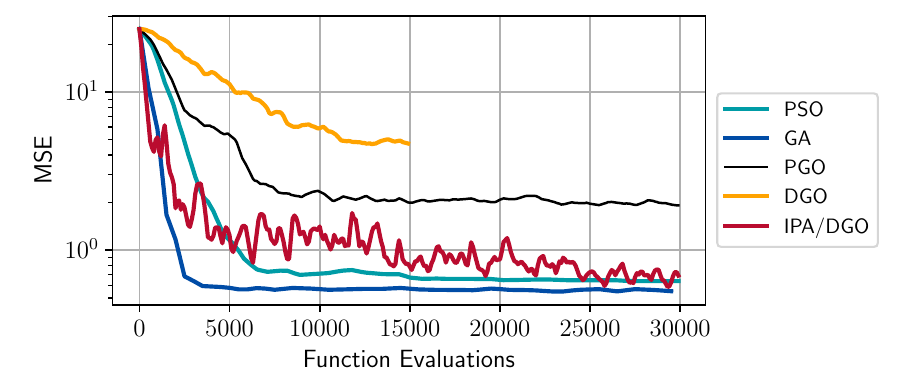}}
\caption{Calibration progress over steps and wall time during calibration of the number of evacuations via the Social Force weights.}
\label{fig:opt_progress_num_evac}
\end{figure*}

For each estimator, PSO, and GA, we select the configuration that produced the best average solution at the end of the time budget of 5 minutes wall time and report the mean over macroreplications of the crisp simulation output, i.e., without any perturbations, which we evaluated in a post-processing step over 1\,000 microreplications.
In addition to the progress over wall time, we also show the progress over function evaluations, each of which reflects one simulation trajectory executed by a sample (gradient estimators), particle (PSO), or population member (GA).

Figure~\ref{fig:opt_progress_y_pos} shows the solution quality over function evaluations and wall time for the first problem.
As expected, we observe that all approaches converge quickly for this problem.
The fastest progress over time is achieved by GA, which converged after only a few seconds.
Considering the gradient estimators, DGO and the finite differences-based PGO exhibit comparatively high variance and are the slowest to converge.
DGO's progress per function evaluation is similar to GA, but is slowed in wall time by the gradient estimation overhead.
Remarkably, IPA, which fared worst in terms of gradient fidelity (cf.~Figure~\ref{fig:fidelity_heatmap_y_pos}), yields the fastest convergence, which suggests that capturing the general curvature of the objective function suffices to quickly identify a local minimum.

In the second calibration problem shown in Figure~\ref{fig:opt_progress_num_evac}, GA outperforms the other methods both in terms of function evaluations and time.
Of the gradient estimators, PGO makes the fastest progress in wall time, with a plateau between about 10s to 100s stemming from a lack of initial progress in one of the 20 macroreplications.
Again, the AD-based gradient estimation benefits from disregarding jumps in the force calculations, which allows IPA/DGO to overtake PGO over function evaluations, albeit encumbered by the AD overhead.

\begin{figure*}[t!]
\subfloat[Progress over wall time.]{\includegraphics[height=2.5cm]{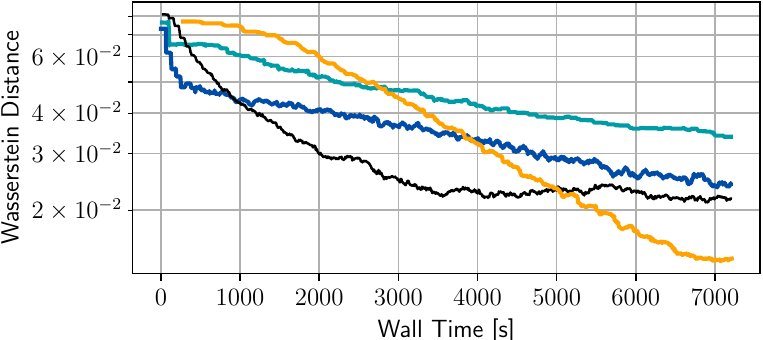}}\hfill
\subfloat[Progress over steps.]{\includegraphics[height=2.5cm]{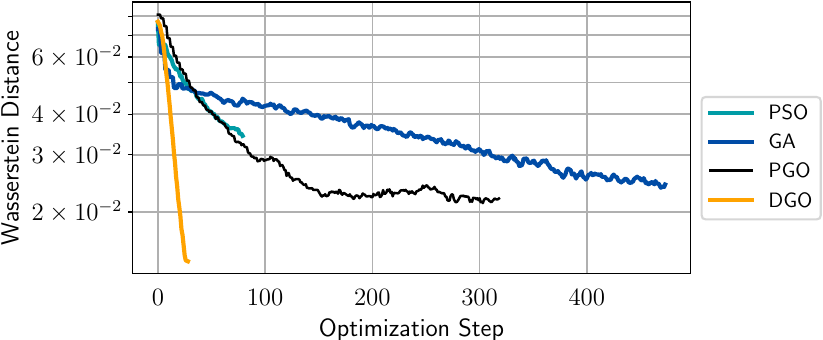}}
\caption{Calibration progress over optimization steps, function evaluations, and wall time for the exit selection scenario. Gradient descent using the sampling-based estimators PGO and DGO achieved the best fit to the reference.}
\label{fig:opt_selection}
\vspace{0.3cm}
\end{figure*}

\subsection{Calibrating Decision-Making Parameters}

Next, we turn to the higher-dimensional problem of adjusting a distribution of weight coefficients that govern the agents' discrete exit selection decisions.
An important property regarding the gradient estimation and a key difference to the previous problems lies in drawing the individual agents' coefficients from the input distribution as part of a simulation run.
Each coefficient is determined by inverse transform sampling on the distribution specified in discretized form by the parameters.
After drawing $u \sim U(0, 1)$, we iterate over the histogram's normalized per-bin probabilities $b_i$ and select the coefficient according to the lowest bin index $i$ with $u \ge \sum_{i=1}^{20} b_i$.
Drawing each pedestrian's coefficient thus involves a sequence of conditional branches on the cumulative sum of normalized bin weights.
The branch conditions are functions of the input parameters and are thus taken into account as part of DGO's gradient estimation.
However, the remainder of each simulation trajectory, including all force calculations, exit selection decisions, and the calculation of the distance to the reference output distribution, is a direct consequence of the branches taken while drawing coefficients.
Since the pathwise derivatives wrt.~parameters thus extend only to the initial coefficient calculation, DGO's gradient estimates are based solely on the critical events generated by these initial branches.
Hence, AD-based derivatives across the force calculations do not contribute to DGO's estimates, which eliminates the main source of noise observed in the previous problems.

Figure~\ref{fig:opt_selection} shows the calibration progress for this problem based on a subset of the hyperparameters from Table~\ref{tab:hyperparams}, and setting the number of microreplications to 1 or 10.
As one function evaluation corresponds to about 1s of wall time, we omit the progress over function evaluations and instead show the progress over optimization steps, each of which can cover several evaluations depending on the method's hyperparameters.
Here, gradient descent using PGO and DGO identified the best solutions within the time budget.
We note the steepness of DGO's progress over steps in the best-performing configuration of 10 simulation replications and 25 samples per step, in contrast to PGO's 1 replication and 25 samples.
DGO overtakes PGO at around 5\,000 steps, whereas PGO stagnates.
PSO and GA show similar progress, and neither reaches convergence by the end of the time budget, with GA achieving a similar solution quality as PGO.

\begin{figure*}[t!]
\centering
\subfloat[Input Distribution over Exit Selection Coefficients]{\includegraphics[width=0.45\textwidth]{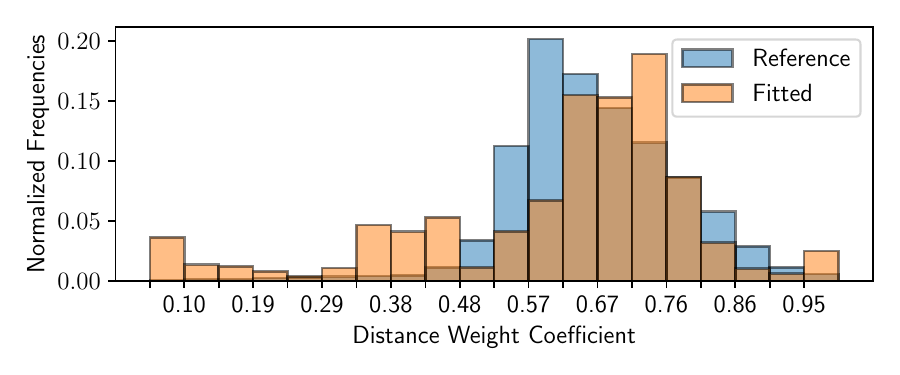}}\hfill
\subfloat[Output Distribution over Evacuation Times]{\includegraphics[width=0.45\textwidth]{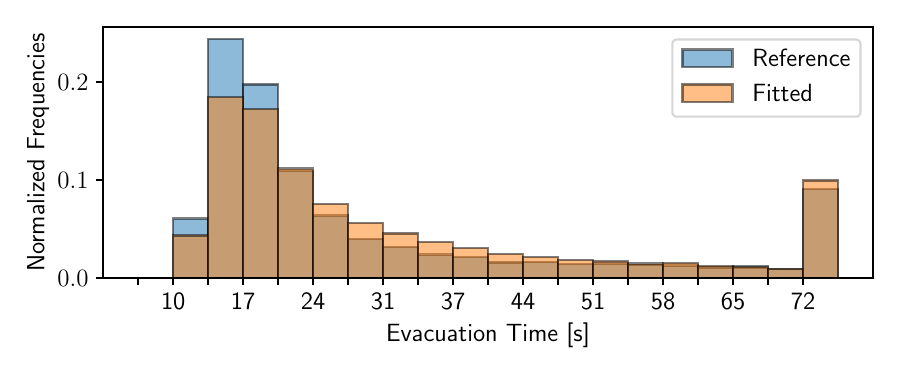}}
\caption{Best calibration results generated by DGO. While the output matches the reference distribution well, the identified input distribution is similar in mean to the reference but differs in shape.}
\label{fig:best_input_output_histograms}
\end{figure*}

Finally, we consider the solution identified by DGO in its best-performing hyperparameter combination and macroreplication.
Figure~\ref{fig:best_input_output_histograms} shows the calibrated input histogram over exit selection coefficients, and the output histogram over evacuation times.
As expected, a good fit is achieved with respect to the distribution of evacuation times.
In contrast, the input distribution still somewhat deviates from the reference in shape, although the tendency is captured.
This result shows that for this problem, high-quality solutions can be achieved via differently shaped input distributions.
Hence, the calibration for real-world purposes would likely benefit from additional criteria to increase identifiability.

\section{Conclusions}
\label{sec:conclusions}

Our study of the crowd model calibration via gradient descent using sampling-based estimators shows both the challenges and the promise of the approach.

Firstly, even in the absence of explicit conditional branching, crowd simulations based on Social Force can observe large jumps in the simulation output.
While our AD-based gradient estimator is capable of accounting for such jumps, scenarios of non-trivial size can generate sufficiently rugged output so that capturing the jumps based on only few samples becomes challenging.
The general function curvature seems to be largely independent of the jumps, which suggests using either simple pathwise derivatives for the Social Force portion of a simulation, or resorting to black-box gradient estimators.

Our largest calibration problem involved adjusting an input distribution across per-agent coefficients, from which we draw by inverse transform sampling using a series of conditional branches.
A key insight is that in this formulation of simulation-based inference, our AD-based method reduces to estimating these branches' effects.
This can positively affect both performance, by reducing the AD overhead to a minimum, and fidelity, by reducing the dependence on potentially noisy intermediate pathwise gradients.
We consider this an encouraging result demonstrating a gainful integration of AD into inference workflows \cite{cranmer2020frontier}.

In a calibration problem across 20 input dimensions, gradient descent using the sampling-based estimators outperformed the gradient-free methods.
The general tendency observed is that the AD-based gradient estimators are beneficial where large jumps in the underlying force calculations can be disregarded and only explicit conditional branches must be accounted for.
Thus, a promising direction for future work lies in further facilitating gradient-based calibration and optimization by model refinements that reduce jumps in acceleration~\cite{kreiss2021deep} while maintaining a realistic representation of real-world crowd behavior.

\section*{Acknowledgements}
Funded by the Deutsche Forschungsgemeinschaft (DFG, German Research Foundation), grant no. 497901036 (PA) and 320435134 (JK).

\bibliographystyle{unsrt}
\bibliography{references.bib}

\begin{thebibliography}{10}

\bibitem{motieyan2018agent}
Hamid Motieyan and Mohammad~Saadi Mesgari.
\newblock An agent-based modeling approach for sustainable urban planning from
  land use and public transit perspectives.
\newblock {\em Cities}, 81:91--100, 2018.

\bibitem{gonzalez2021agent}
Mauricio Gonz{\'a}lez-M{\'e}ndez, Camilo Olaya, Isidoro Fasolino, Michele
  Grimaldi, and Nelson Obreg{\'o}n.
\newblock Agent-based modeling for urban development planning based on human
  needs. conceptual basis and model formulation.
\newblock {\em Land Use Policy}, 101:105110, 2021.

\bibitem{hackl2019epidemic}
J{\"u}rgen Hackl and Thibaut Dubernet.
\newblock Epidemic spreading in urban areas using agent-based transportation
  models.
\newblock {\em Future internet}, 11(4):92, 2019.

\bibitem{chen2008agent}
Xuwei Chen and F~Benjamin Zhan.
\newblock Agent-based modelling and simulation of urban evacuation: relative
  effectiveness of simultaneous and staged evacuation strategies.
\newblock {\em Journal of the Operational Research Society}, 59(1):25--33,
  2008.

\bibitem{yin2014agent}
Weihao Yin, Pamela Murray-Tuite, Satish~V Ukkusuri, and Hugh Gladwin.
\newblock An agent-based modeling system for travel demand simulation for
  hurricane evacuation.
\newblock {\em Transportation research part C: emerging technologies},
  42:44--59, 2014.

\bibitem{kasereka2018agent}
Selain Kasereka, Nathana{\"e}l Kasoro, Kyandoghere Kyamakya,
  Emile-Franc~Doungmo Goufo, Abiola~P Chokki, and Maurice~V Yengo.
\newblock Agent-based modelling and simulation for evacuation of people from a
  building in case of fire.
\newblock {\em Procedia Computer Science}, 130:10--17, 2018.

\bibitem{wolinski2014parameter}
David Wolinski, S~J.~Guy, A-H Olivier, Ming Lin, Dinesh Manocha, and Julien
  Pettr{\'e}.
\newblock Parameter estimation and comparative evaluation of crowd simulations.
\newblock In {\em Computer Graphics Forum}, volume~33, pages 303--312. Wiley
  Online Library, 2014.

\bibitem{voloshin2015optimization}
Daniil Voloshin, Dmitriy Rybokonenko, and Vladislav Karbovskii.
\newblock Optimization-based calibration for micro-level agent-based simulation
  of pedestrian behavior in public spaces.
\newblock {\em Procedia Computer Science}, 66:372--381, 2015.

\bibitem{pietzsch2020metamodels}
Bruno Pietzsch, Sebastian Fiedler, Kai~G Mertens, Markus Richter, C{\'e}dric
  Scherer, Kirana Widyastuti, Marie-Christin Wimmler, Liubov Zakharova, and Uta
  Berger.
\newblock Metamodels for evaluating, calibrating and applying agent-based
  models: a review.
\newblock {\em The Journal of academic social science studies}, 23(2), 2020.

\bibitem{bode2020parameter}
Nikolai Bode.
\newblock Parameter calibration in crowd simulation models using approximate
  bayesian computation.
\newblock {\em arXiv preprint arXiv:2001.10330}, 2020.

\bibitem{godel2022bayesian}
Marion G{\"o}del, Nikolai Bode, Gerta K{\"o}ster, and Hans-Joachim Bungartz.
\newblock Bayesian inference methods to calibrate crowd dynamics models for
  safety applications.
\newblock {\em Safety science}, 147:105586, 2022.

\bibitem{cranmer2020frontier}
Kyle Cranmer, Johann Brehmer, and Gilles Louppe.
\newblock The frontier of simulation-based inference.
\newblock {\em Proceedings of the National Academy of Sciences},
  117(48):30055--30062, 2020.

\bibitem{griewank2008evaluating}
Andreas Griewank and Andrea Walther.
\newblock {\em Evaluating derivatives: principles and techniques of algorithmic
  differentiation}.
\newblock SIAM, 2008.

\bibitem{margossian2019review}
Charles~C Margossian.
\newblock A review of automatic differentiation and its efficient
  implementation.
\newblock {\em Wiley Interdisciplinary Reviews: Data Mining and Knowledge
  Discovery}, 9(4):e1305, 2019.

\bibitem{gong1987smoothed}
Wei-Bo Gong and Yu-Chi Ho.
\newblock Smoothed (conditional) perturbation analysis of discrete event
  dynamical systems.
\newblock {\em IEEE Transactions on Automatic Control}, 32(10):858--866, 1987.

\bibitem{chopra2021deepabm}
Ayush Chopra, Ramesh Raskar, Jayakumar Subramanian, Balaji Krishnamurthy,
  Esma~S Gel, Santiago Romero-Brufau, Kalyan~S Pasupathy, and Thomas~C
  Kingsley.
\newblock Deepabm: scalable and efficient agent-based simulations via geometric
  learning frameworks - a case study for covid-19 spread and interventions.
\newblock In {\em Winter Simulation Conference (WSC)}, pages 1--12. IEEE, 2021.

\bibitem{son2022differentiable}
Sanghyun Son, Yi-Ling Qiao, Jason Sewall, and Ming~C Lin.
\newblock Differentiable hybrid traffic simulation.
\newblock {\em ACM Transactions on Graphics (TOG)}, 41(6):1--10, 2022.

\bibitem{andelfinger2023towards}
Philipp Andelfinger.
\newblock Towards differentiable agent-based simulation.
\newblock {\em ACM Transactions on Modeling and Computer Simulation},
  32(4):1--26, 2023.

\bibitem{arya2022automatic}
Gaurav Arya, Moritz Schauer, Frank Sch\"{a}fer, and Christopher Rackauckas.
\newblock Automatic differentiation of programs with discrete randomness.
\newblock In S.~Koyejo, S.~Mohamed, A.~Agarwal, D.~Belgrave, K.~Cho, and A.~Oh,
  editors, {\em Advances in Neural Information Processing Systems}, volume~35,
  pages 10435--10447. Curran Associates, Inc., 2022.

\bibitem{kreikemeyer2023smoothing}
Justin~N. Kreikemeyer and Philipp Andelfinger.
\newblock Smoothing methods for automatic differentiation across conditional
  branches.
\newblock {\em IEEE Access}, 11:143190--143211, 2023.

\bibitem{nesterov2017random}
Yurii Nesterov and Vladimir Spokoiny.
\newblock Random gradient-free minimization of convex functions.
\newblock {\em Foundations of Computational Mathematics}, 17:527--566, 2017.

\bibitem{helbing1995social}
Dirk Helbing and Peter Molnar.
\newblock Social force model for pedestrian dynamics.
\newblock {\em Physical review E}, 51(5):4282, 1995.

\bibitem{seyer2023differentiable}
Ruben Seyer.
\newblock Differentiable monte carlo samplers with piecewise deterministic
  markov processes.
\newblock Master's thesis, Chalmers University of Technology, 2023.

\bibitem{fu2006chapter}
Michael~C. Fu.
\newblock {Chapter 19: Gradient Estimation}.
\newblock In Shane~G. Henderson and Barry~L. Nelson, editors, {\em Simulation},
  volume~13 of {\em Handbooks in Operations Research and Management Science},
  pages 575--616. Elsevier, 2006.

\bibitem{ho1983new}
Yu-Chi Ho and Christos Cassandras.
\newblock A new approach to the analysis of discrete event dynamic systems.
\newblock {\em Automatica}, 19(2):149--167, 1983.

\bibitem{christodoulou2023differentiable}
Sebastian Christodoulou and Uwe Naumann.
\newblock Differentiable programming: Efficient smoothing of
  control-flow-induced discontinuities.
\newblock {\em arXiv preprint arXiv:2305.06692}, 2023.

\bibitem{chaudhuri2010smooth}
Swarat Chaudhuri and Armando Solar-Lezama.
\newblock Smooth interpretation.
\newblock {\em ACM Sigplan Notices}, 45(6):279--291, 2010.

\bibitem{williams1992simple}
Ronald~J Williams.
\newblock Simple statistical gradient-following algorithms for connectionist
  reinforcement learning.
\newblock {\em Machine learning}, 8:229--256, 1992.

\bibitem{polyak1987introduction}
B.T. Polyak.
\newblock {\em Introduction to Optimization}.
\newblock Optimization Software, New York, 1987.

\bibitem{wang2022modeling}
Xiang Wang, Chraibi Mohcine, Juan Chen, Ruoyu Li, and Jian Ma.
\newblock Modeling boundedly rational route choice in crowd evacuation
  processes.
\newblock {\em Safety Science}, 147:105590, 2022.

\bibitem{miranda2018pyswarms}
Lester~James Miranda.
\newblock Pyswarms: a research toolkit for particle swarm optimization in
  python.
\newblock {\em Journal of Open Source Software}, 3(21):433, 2018.

\bibitem{kreiss2021deep}
Sven Kreiss.
\newblock Deep social force.
\newblock {\em arXiv preprint arXiv:2109.12081}, 2021.

\end{thebibliography}
\end{document}